# Overview: Computer vision and machine learning for microstructural characterization and analysis

Elizabeth A. Holm[1*], Ryan Cohn[1], Nan Gao[1], Andrew R. Kitahara[1], Thomas P. Matson[1,2], Bo Lei[1], Srujana Rao Yarasi[1]

[1]Department of Materials Science and Engineering, Carnegie Mellon University, Pittsburgh, PA 15213 USA

[2]Department of Materials Science and Engineering, MIT, Cambridge, MA 02139 USA

## ABSTRACT

The characterization and analysis of microstructure is the foundation of microstructural science, connecting the materials structure to its composition, process history, and properties. Microstructural quantification traditionally involves a human deciding *a priori* what to measure and then devising a purpose-built method for doing so. However, recent advances in data science, including computer vision (CV) and machine learning (ML) offer new approaches to extracting information from microstructural images. This overview surveys CV approaches to numerically encode the visual information contained in a microstructural image, which then provides input to supervised or unsupervised ML algorithms that find associations and trends in the high-dimensional image representation. CV/ML systems for microstructural characterization and analysis span the taxonomy of image analysis tasks, including image classification, semantic segmentation, object detection, and instance segmentation. These tools enable new approaches to microstructural analysis, including the development of new, rich visual metrics and the discovery of processing-microstructure-property relationships.

---

[*] Corresponding author email: eaholm@andrew.cmu.edu

# 1 INTRODUCTION: THE QUANTIFICATION OF MICROSTRUCTURE

In 1863, the geologist Henry Clifton Sorby examined acid-etched and polished steel under a microscope and observed a complex collection of substructures that we now call microstructure.[1] Over the next two decades, Sorby related these visual entities to the chemistry, history, and behavior of various steel alloys, making the first connections between materials structure, composition, processing, and properties.[2] Sorby's observations were necessarily qualitative; for example, he wrote, "There is also often great variation in the size of the crystals [grains] of iron…and one cannot but suspect that such great irregularities might be the cause of the fracture..."[2] However, by the early 1900's, methods for measuring microstructural features had been developed, and in 1916 the first ASTM metallographic standard E 2-17T included planimetric grain size measurement.[3]

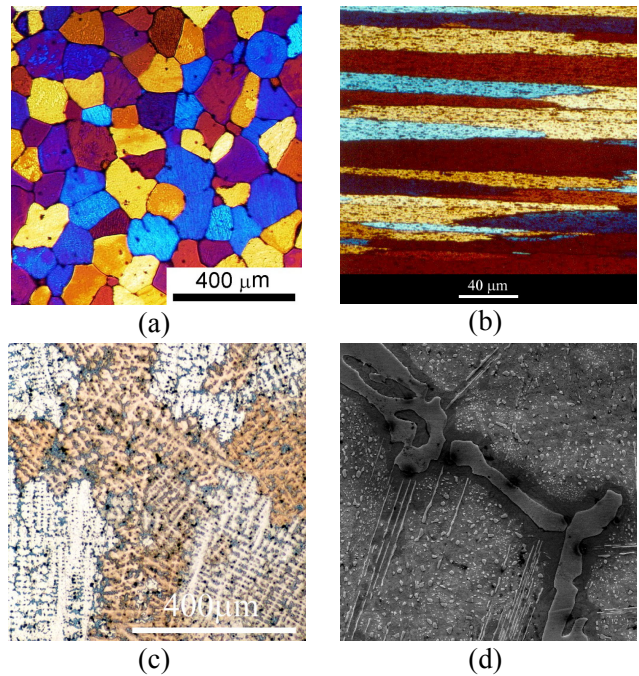

Figure 1. Microstructural diversity. (a) Equiaxed grains in an Al alloy seen in polarized light.[4] (b) Elongated grains in extruded Al, polarized light micrograph.[4] (c) Dendritic colonies in cast bronze, true color light micrograph.[4] (d) Carbide structure in ultrahigh carbon steel, SEM image.[5]

Throughout the 20$^{th}$ century, metallurgists and materials scientists continued to create and refine an ever-growing catalog of microstructural metrics.[6] The field of quantitative stereology linked two-dimensional (2D) cross-sectional structures to their true three-dimensional (3D) counterparts,[7] techniques such as electron backscatter diffraction (EBSD) microscopy were especially amenable to digital analysis, and with the advent of 3D x-ray diffraction microscopy (3DXRD),[8] complete 3D microstructural representations became a reality.[9] Despite the sustained progress in microstructural science, however, an overarching microstructural metric remained elusive. Instead, each individual metric was developed and is applied on a case-by-case basis. The reason for this is simple: The visual phase space of microstructural images is immense, as illustrated in Figure 1. This leads to the two fundamental problems of quantitative microstructural science:

1. ***What to measure.*** For any given microstructure, it is not always self-evident which metric is most strongly related to the process or property of interest. Consider the Hall-Petch relationship $\sigma_y = kd^{1/2}$, which directly links average grain diameter $d$ with yield strength $\sigma_y$ via a proportionality constant $k$.[10] In an equiaxed, single-phase polycrystal such as Figure 1(a), it is a simple matter to measure $d$.

But if the grains are elongated as in Figure 1(b), is *d* still the relevant metric, or is the grain aspect ratio more important?

As microstructures become more complex, so does metric selection. Which is the most influential scale in Figure 1(c): dendrite arm spacing, dendrite colony size, or porosity distribution? Of course, there is the very real possibility that multiple metrics may arise from a given process history or contribute to a given property. Ultimately, deciding what microstructural feature(s) to measure requires human judgment informed by knowledge and experience.

2. *How to measure it.* While some microstructural metrics are relatively straightforward to measure, others are not. Consider the steel microstructure in Figure 1(d). A feature that affects this alloy's mechanical properties is the size of the carbon denuded zone (the dark area bordering the network carbides). Although it is simple in principle to measure its average dimension, it requires locating the boundaries of the denuded zone via segmentation. (Segmentation is defined as determining feature membership for each pixel in an image). Because the denuded zone boundary is fuzzy, segmentation is problematic and usually is left to the subjective judgment of a human expert. The outcome can be uncertain or even irreproducible results.

Segmentation issues may affect even apparently uncomplicated measurements. For example, in polycrystalline microstructures like Figure 1(a), some grain boundaries may not etch deeply enough to create a dark line in an optical image. While the human eye is able to locate these "missing boundaries" based on visual clues, automating this process has remained stubbornly intractable.[11] Thus, even the simple metric of grain size can harbor complex measurement challenges.

Even now, 150 years after Sorby noted the iridescent sheen of "the pearly constituent" (pearlite), microstructural analysis often relies on qualitative, rather than quantitative, descriptions. Terms like equiaxed, aligned, rough, dispersed, ordered, columnar, etc. are often used without quantification.

Computer vision (CV) is the computer science field that focuses on quantifying the visual information content of digital images.[12] A digital image is a numeric representation where each pixel has an integer gray value or a short vector color value (i.e. RGB) associated with it; the individual pixels provide little information about the content of the image. The goal of CV is to aggregate pixels to represent an image as a high-dimensional tensor of visual information. While CV may be familiar from facial recognition and self-driving cars, it is also used in applications that are strikingly similar to some microstructural quantification tasks, such as determining how many people appear in a picture of a crowd[13] or what proportion of a satellite photograph is comprised of farmland.[14]

In order to extract abstract information from the image representation produced by a CV scheme, we can use methods for finding correlations in high-dimensional data space, many of which fall under the general classification of machine learning (ML).[15] ML approaches may be supervised or unsupervised. Supervised ML involves training a system based on human-determined ground-truths. For example, given a set of photographs with metadata noting the presence or absence of a cat, a supervised ML system can learn to identify images of cats. Unsupervised learning algorithms find relationships between image representations without ground-truth data or human intervention, typically by generating clusters of related images. These approaches are complementary, and each is applicable to different problems.

A number of recent studies have applied CV and ML approaches to develop a more general approach to quantitative microstructural analysis.[16] This is a promising direction for creating tools that capture rich and complete microstructural information (what to measure) in a quantitative, objective, and general manner (how to measure it). In this paper, we present applications of CV and ML for a variety of microstructural image analysis tasks.[17] We show that the CV/ML approach can assist, improve, or even replace traditional, *ad hoc* microstructural characterization methods.

## 2 METHODS

Accurate microstructural measurements are fundamental to making structure-processing-property connections. However, there are many microstructural analysis tasks that require human judgment and so have an element of subjectivity that makes them difficult to automate and susceptible to bias. Because artificial intelligence methods, including CV and ML, can learn to replicate human visual judgments, these tools are good candidates to perform these tasks in an objective, autonomous, and efficient manner.

The fields of CV and ML are rapidly evolving. There is no reason to believe that the particular methods presented here are optimal or even will remain competitive as new methods are developed. However, we expect that the concepts and approaches of CV and ML will continue to be valuable tools for microstructural science. For that reason, this section will present the foundational bases of CV and ML, which are common across methods. Details of the specific computational codes used in the examples are available in the supplementary material, Table SI.

### 2.1 Computer Vision: Create a numerical representation

Computer vision encompasses an array of methods for creating a numerical representation of a visual image, termed the feature vector; most of these methods are optimized for specific tasks (facial recognition, object identification, texture analysis, etc.).[12, 18] Across application domains, however, there are two basic approaches to CV: feature-based representations and representations based on convolutional neural networks (CNNs). Feature-based methods create an image representation that is in essence a statistical representation of the visual features in the image.[17b, 17c, 17e, 17f, 18s, 19] Filters that activate when they encounter a feature[18h, 18k] (typically an edge, corner, or blob) are applied to the original image. Each feature is then numerically encoded with a descriptor,[18e, 18k] and the image representation is some aggregation of the feature descriptors.[20] CNNs also use filter activations as visual features.[21] The primary difference between feature-based and CNN image representations is that while the filters for feature-based methods are selected by human experts, the CNN filters are learned during the training and optimization of the CNN. Moreover, in typical CNN tasks, the filter activations are not abstracted into an image representation; instead, they are used directly as the feature map for the image.[22] The result is that, since the advent of the AlexNet CNN in 2012,[23] CNNs have outperformed feature-based methods (and often human experts[24]) for essentially all CV tasks.[25] Thus, we will restrict our discussion to CNN-based feature vectors.

The fundamental objective of CV is to represent the visual content of an image in numerical form, and there are numerous methods to accomplish this via CNNs. We will focus on two methods that are particularly suitable for microstructural images: CNN layers and hypercolumn pixels.

Convolutional neural networks take an image (or image-like data) as input, apply a variety of signal processing operations to it in order to encode it as a vector, and then utilize an artificial neural network[25-26] or other ML method to draw a conclusion about the visual content of the image. The first part of the CNN pipeline – encoding the image as a feature vector – is termed the feature learning stage, and the second part – drawing a conclusion – is the classification stage.

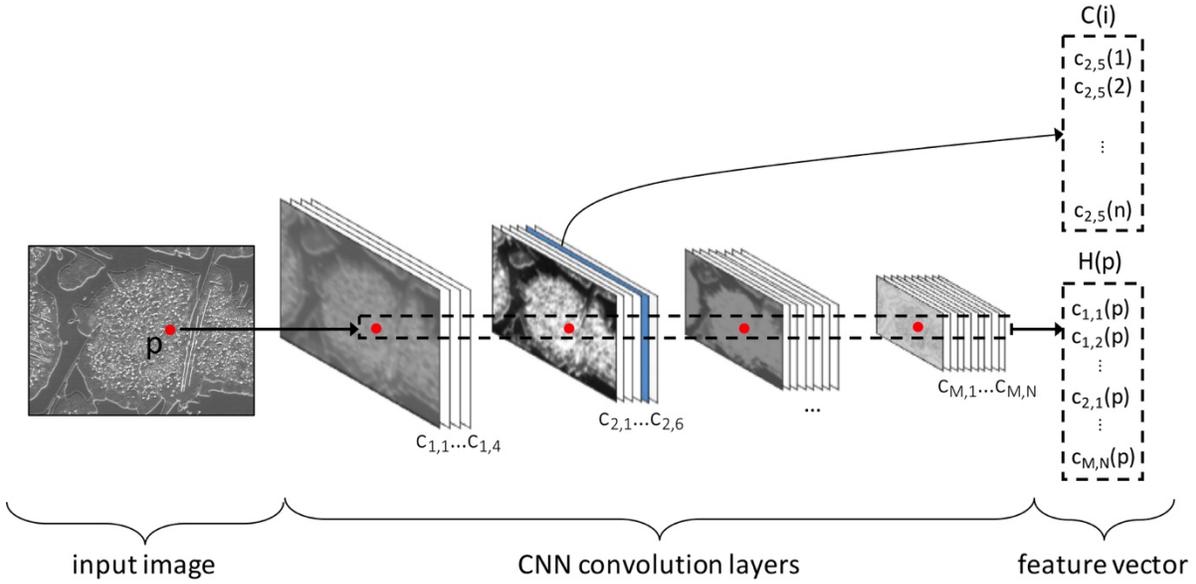

*Figure 2. The feature learning stage of a CNN. The input image i (left) is processed through layers $c_{1,1}$ through $c_{M,N}$. Pooling decreases image resolution between groups (or blocks) of layers. A feature vector for the image as a whole, C(i), may be constructed by flattening any of the layers (here, the $c_{2,5}$ convolution layer shown in blue). Alternatively, a feature vector for each pixel p may be assembled from filter activations in all the layers, giving the hypercolumn pixel vector, H(p).*

In the feature learning stage, shown in Figure 2, the CNN rasters (convolves) sets of filter patches pixel-by-pixel and records the filter activation values. It does this hierarchically, first to the image (generating the first convolution layer) and then to the subsequent layers (generating the second through $n^{th}$ convolution layers). After several convolutions, activations are rectified (typically via a rectified linear unit, ReLU, filter that converts negative values to zero). Pooling (downsampling) then combines multiple pixels in one layer into a single pixel in the next layer, and another set of convolutions are performed. After a number of convolution and pooling iterations, the final layer (the fully connected layer) is flattened (written as a vector), forming the representation used for decision-making in the classification stage.[22, 27]

Designing and training a CNN requires deep expertise and a large data set, making it impractical for most microstructural data sets. However, CNNs that have been optimized and trained on a large set of natural images have been successfully used with other kinds of images, including microstructures. This transferability of results is likely due to the fact that images of very different things share common visual features, including edges, blobs, and visual textures. Thus, transfer learning[20a] enables us to use pre-trained CNNs (such as the VGG16 network[22] trained on the ImageNet data set[28]) for microstructural representation. However, since we are not interested in classifying microstructural images into the ImageNet categories (broccoli, bucket, bassoon,…), we truncate the network before the classification stage. Instead, we use the CNN layers themselves as the image representations for ML tasks, as shown in Figure 2, often using a dimensionality-reducing encoding such as principal component analysis (PCA)[29] or vector of locally aggregated descriptors (VLAD)[20d] to decrease the length of the feature vector for efficient computing.[17b]

Which layer of a CNN representation is the best choice for a feature vector depends on the characteristics of the micrographs.[30] Because pooling operations cause pixels in deep layers to represent large areas of the original image, deep layers capture features at large length scales. Conversely, filter activations in shallow layers represent local environments. Thus, visually simple micrographs may be better represented by shallow layers, and complex microstructures may be better represented by deep layers, which capture multiscale structures more completely. However, there is not yet a mechanism for determining the optimum layer to represent a given type of microstructure *a priori*; instead, the decision is made by trial and error.[17b, 30]

For a more complete image representation, we might wish to combine the information from all of the convolution layers; this is the basis for the hypercolumn pixel feature vector,[18b] as shown in Figure 2. We begin with an image and its convolution layers in a pre-trained CNN. For each of the selected pixels in the original image, the hypercolumn is built by stacking the activations of the convolution layers at the pixel's real-space position; the result is an image representation where each hypercolumn stores information about the pixel's feature membership at multiple length scales. Representing an image by the hypercolumns of every pixel is memory intensive, so typically we select a sparse subset of pixels. Hypercolumn features are primarily used for image segmentation tasks, which we discuss below.[31]

## 2.2 Machine Learning: Extract quantitative visual information

The goal of ML is to extract quantitative visual information from the high-dimensional feature vector. This information might comprise a classification (e.g. ferritic, austenitic, martensitic,…), an association with a metadata value (e.g. yield strength), a measurement (e.g. grain size), presence of a particular feature (e.g. surface defect), or any other value that might be contained in the feature vector. ML methods are either supervised (trained using known correct answers, termed ground truth) or unsupervised (finding patterns without knowledge of a ground truth), and there are important roles for each approach.

There is a wide array of supervised ML methods,[25-26, 32] and the choice of method depends on the application. Supervised ML systems that are widely used for visual image feature vector data include support vectors machines (SVM),[33] random forest (RF)[34] classifiers, and deep learning methods such as artificial neural networks (ANNs).[22, 25-26, 27] SVMs operate by learning the set of hyperplanes that best separates feature vectors into groups according to their ground truth type or class. Once the separating planes are known, additional vectors can be associated with the appropriate group, i.e. classified. Advantages of SVM are flexibility, generality, and performance. However, the success of an SVM model depends on whether the high-dimensional data structure is amenable to planar separation.

RF classifiers begin by constructing a set of decision trees to predict the class of an image; within each tree, decisions are based on values of feature vector elements. To classify the image, the trees each offer a prediction, and majority rules. In the training phase of an RF classifier, the feature vector elements that populate the trees and their decision values are optimized to give the best match with the known ground truth. Once the trees are optimized, they can be applied to classify additional vectors. An advantage of RFs is interpretability, since the basis for decision is easily verified.[35] However, RFs are not always easy to apply to complex image representations.

ANNs process the feature vector through the hidden layers of a neural network[32] in order to make a prediction regarding the image. The structure and connectivity of the ANN can take many forms, and its architecture is selected to maximize performance, often by trial and error. During the learning process, the weights of the connections between neurons in the input, hidden, and output layers are optimized to give the best match with the known ground truth. Once the weights have been determined, the ANN can make predictions about previously unseen vectors. ANNs as a class are immensely flexible and scale to handle both large amounts of data and very high dimensional data. However, they are black box models, where the basis for decision can be difficult to parse.

Unsupervised ML algorithms find relationships between image representations without ground-truth data or human intervention, typically by generating clusters of related images. *k*-means is one example of an unsupervised clustering method.[36] For a set of *N*-dimensional feature vectors, *k*-means groups them into a user-specified number of *N*-dimensional clusters that minimize a cost function that is some measure of "cluster goodness," such as cluster compactness. Finding a globally optimal clustering is an np hard computational problem (i.e. requires exponential computing time), so *k*-means uses various computing strategies to find good solutions. An implication of this is that for a given set of vectors, *k*-means results may vary depending on computational parameters. An advantage of *k*-means clustering is that because it identifies a set of cluster centroids, additional vectors can be associated with clusters straightforwardly. Thus, it can be used as a basis for classification. A disadvantage is that, as a full *N*-dimensional representation, it may not be easy to visualize the results in 2D or 3D.

A powerful tool for reducing high-dimensional data to lower-dimensional clusters is t-distributed stochastic neighbor embedding (t-SNE).[37] t-SNE weights image similarity on a nonlinear scale that diminishes quickly as image similarity decreases; thus t-SNE highly favors grouping similar images but does not capture relationships between dissimilar images. An advantage of t-SNE is that it often is able to resolve low-dimensional clusters suitable for visualization and analysis. A disadvantage is that since t-SNE is constructed based on pairwise comparisons of feature vectors, it is not possible to add additional data to a t-SNE map without recomputing it entirely, so t-SNE cannot be used for classification.

The choice of ML modality and model depends on the nature of the input data and the desired outcome. Often, multiple approaches are attempted and evaluated for performance and/or other advantages and disadvantages. In this process, it is helpful to include a domain expert in ML algorithms, since the best-in-class solutions are ever-evolving.

### 2.3 Data: The basis for data science

Perhaps the greatest challenge to the creation of CV/ML approach for microstructural science is the need for microstructural image data suitable for training and utilizing such systems. Data size is usually assumed to be the limiting factor, and in some cases it is. However, as discussed below, we have achieved excellent results in some complex image analysis tasks with very small numbers of training images (sometimes fewer than 10). We believe that this has to do with the data-richness of microstructural images compared to the natural images used in typical CV studies. For instance, where a natural image might contain a cat, a micrograph might contain 100 precipitates. In fact, microscopists strive to capture microstructural images that are statistically representative of the material they portray. Furthermore, the spatial relationships and length scales implicit in a microstructural image are physically-based and meaningful. Finally, the entire image typically constitutes the microstructure; there is no "meaningless" background to subtract or ignore. Thus, microstructural images tend to be rich in relevant information.

Similarly, concerns about image quality are often unfounded. ML methods learn to extract the information contained in the data as it is presented to them; there are no judgments about the quality of focus, resolution, field of view, etc. In fact, exposing the ML system to the full scope of the data space results in a more robust outcome. This is not a recommendation to ignore good microscopy practices, but rather a suggestion not to assume that images must be vetted or weeded out based on human standards of quality.

Data collection practices that do increase the performance of CV/ML systems include image redundancy, standardization, and augmentation. Taking redundant images of a sample with non-overlapping fields of view increases the amount of visual data for a relatively low additional cost, compared to taking one "representative" image. This allows the ML system to learn the full data space, while avoiding overfitting and human bias. Standardizing imaging conditions (such as instrument, settings, magnification, orientation, etc.) and/or performing judicious image preprocessing helps prevent the CV/ML system from learning the wrong thing. That said, there are data sets in which imaging differences are inevitable; in those cases, verifying that the system is learning the relevant information is paramount.[17e] Data augmentation involves manipulating existing images, usually via subsampling, translation, rotation, or affine transformation, to create additional data for training and testing.[38] For example, for micrographs that do not possess a standard orientation, augmenting the original data with rotated versions can more fully sample the visual data space represented by the data set, potentially improving ML outcomes.

The most valuable microstructural data sets include metadata that enriches their information content. Metadata may include material system, composition, imaging information, processing history, property measurements, and any other data available related to the image. Now, more than ever, aggregating and cross-referencing data from multiple sources is the key to discovery.

## 3 RESULTS: A TAXONOMY OF MICROSTRUCTURAL ANALYSIS

From a given image, a microstructural scientist may extract different kinds of information from various visual signals. This information can be arranged in a hierarchy, or taxonomy, of microstructural analysis tasks, which includes:

- Image classification – identifying the content of an image
- Semantic segmentation – associating each pixel in an image with a constituent of the image
- Object detection – locating individual objects in an image
- Instance segmentation – assigning pixels to individual objects

Figure 3 shows this microstructural taxonomy applied to a scanning electron microscopy (SEM) micrograph of metal powder particles. Image classification identifies the image as belonging to the 'powder' class. Semantic segmentation associates pixels with either powder particles or background. Object detection finds each individual powder particle, and instance segmentation assigns pixels to particles. In the following sections, we will demonstrate and discuss the CV and ML applications for tasks in each component of the microstructural analysis taxonomy.

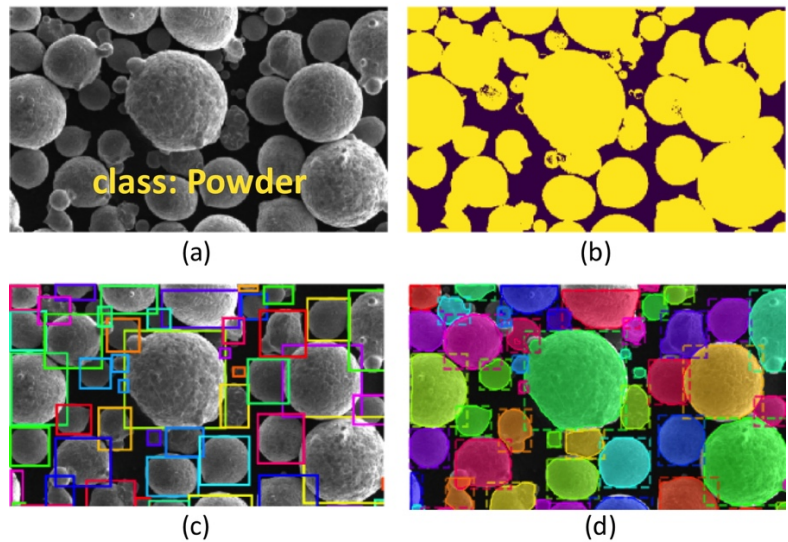

*Figure 3. A taxonomy of microstructural analysis applied to an SEM image of metal powder particles. (a) Image classification identifies the image as belonging to the 'powder' class. (b) Semantic segmentation associates pixels with one of two constituents: powder particles (yellow) or background (purple). (c) Object detection finds each individual powder particle (boxes). (d) Instance segmentation assigns pixels to particular particles (colors). [Original image courtesy of I. Anderson, Ames Lab.]*

### 3.1 Image classification and the feature vector

Image classification may not seem important, since we usually know what our microstructures are. However, classification of images underlies a host of critical archiving and analysis tasks. A CV approach to image classification begins with defining a feature vector, as discussed above.

Since the feature vector numerically encodes the visual information contained in an image, the visual similarity between two images should be related to the numerical similarity between their feature vectors. Thus, a quantitative measure of overall image similarity can be defined by a distance between feature vectors, such as the Euclidean distance ($L^2$ norm). The visual similarity metric then forms the basis for visual search, visual clustering, and classification. For example, feature vectors were computed using a CNN layer representation for a database of 961 microstructures of ultrahigh carbon steels.[5] Figure 4(a) shows the three images with feature vectors closest to that of a given target image; clearly, feature vector similarity is reflected in visual similarity. Notably, while the first two matches contain microconstituents of similar size, fraction, contrast, and structure to the target image, the third is the target image at a lower magnification, demonstrating that the visual search encompasses multiple aspects of visual similarity. Figure 4(b) shows a t-SNE image cluster map where each point represents an image, and the distance between points scales with the distance between feature vectors; point color corresponds to the primary microstructural constituent in each micrograph. Clearly, similar images cluster, which illustrates the visual structure of the data set. Thus, the feature vector enables the primary database functions of search and sort to be performed based on visual, rather than language-based, information.

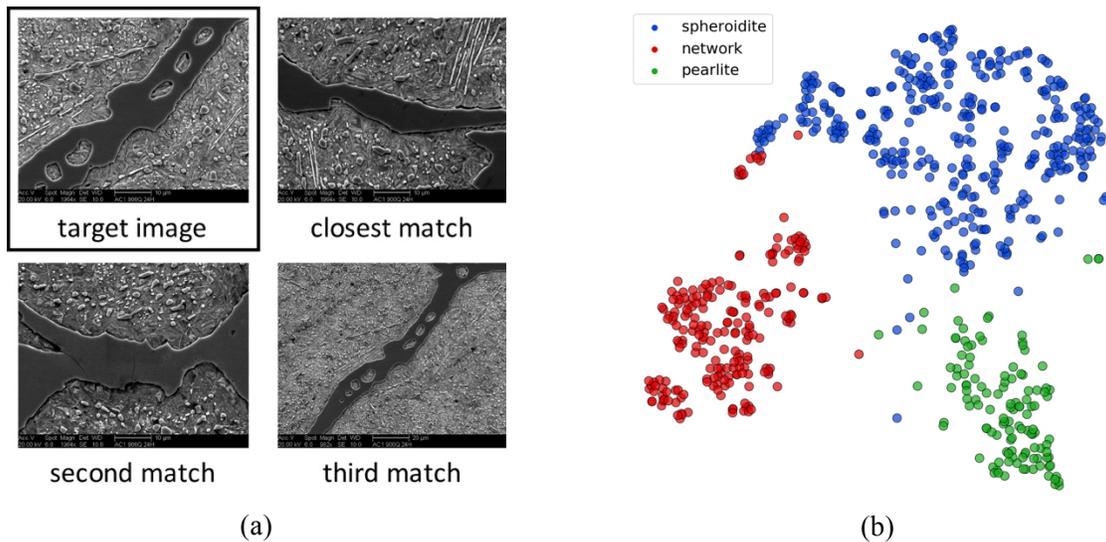

(a) (b)

*Figure 4. Applications of the feature vector for image similarity. (a) Visual search for images similar to the target image in a database of about 900 microstructures of ultrahigh carbon steel.[5] The closest matching images share a number of visual features in common with the target. (b) In this t-SNE plot, micrographs (points) cluster according to their visual similarity, which also corresponds to their primary microconstituent: spheroidite (blue), network carbide (red), or pearlite (green).*

The fact that the visual clusters in Figure 4(b) correspond to microstructural constituents implies that the feature vector can be utilized to detect the presence of specific microstructural features. In fact, a linear support vector machine (SVM) was trained to classify the primary microconstituent in each image in the UHCS database (supervised machine learning), achieving 99±1% accuracy (defined as fraction of correct classifications).[5] Other studies have used unsupervised and supervised ML to detect and classify defects,[39] microstructural consituents,[40] atomic structures,[41] and damage.[42] Thus, image classification contributes to a wide variety of image processing tasks including image analysis, keyword identification, and quality control.

While the feature vector is often used for qualitative tasks, it is worth remembering that it contains information about pixel membership in the various microstructural features and as such can potentially be used to quantify structural metrics directly, without image segmentation. To do so, we need a data set sufficiently large to train an ML system and with unambiguous ground-truth values of the microstructural metric to be measured. For example, to address the problem of missing boundaries in grain size measurement,[11] we generated 15,213 synthetic optical microstructures of pure isotropic polycrystals with known grain sizes, using the SPPARKS kinetic Monte Carlo grain growth simulation code.[43] These were used to train a CNN system[22, 44] with a fully connected regression layer[45] to learn grain size, both for perfect structures and for polycrystals where some of the boundaries were arbitrarily erased.

As shown in Figure 5(a), for perfect polycrystals the system predicts grain size with a standard error of 2.3%; we also found that the standard error increases linearly as the fraction of missing boundaries increases, and reaches 3.9% for a missing boundary fraction of 0.4. These errors are well within an expected measurement fidelity for grain size. For instance, the ASTM standard claims a precision of ±0.5 grain size units for linear intercept methods, which this model meets for all missing boundary fractions (assuming circular grains and converting to a linear grain size metric).[46] The ability to measure without segmentation creates opportunities for fault-tolerant, high-throughput microstructural evaluation.

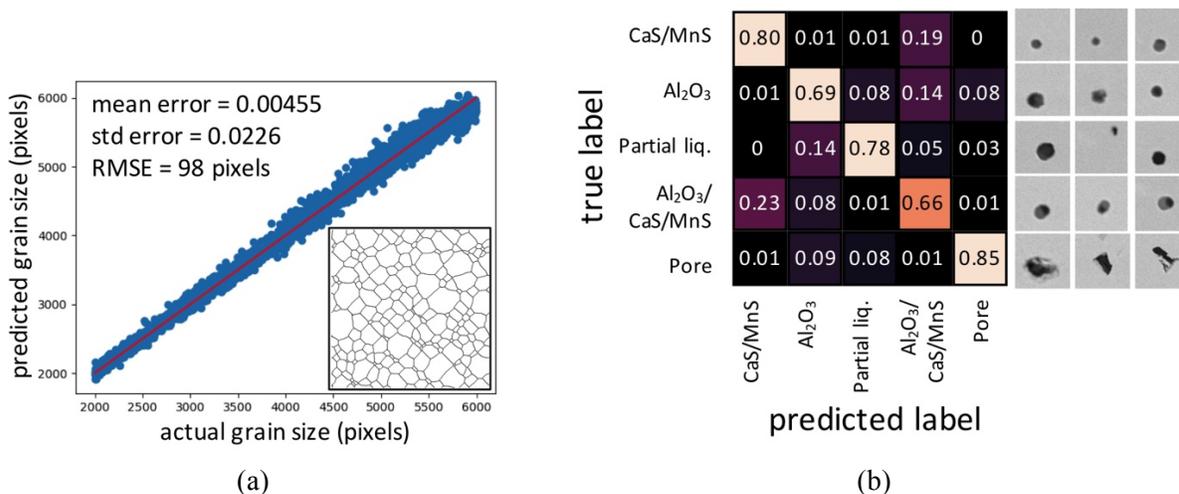

(a) (b)

*Figure 5. Advanced applications of the microstructural feature vector. (a) Determining grain size from simulated polycrystalline micrographs without segmentation or direct measurement using deep regression. Blue points indicate previously unseen test images, and the red line corresponds to perfect accuracy (predicted = actual). Inset shows an example polycrystal. (b) Determining steel inclusion composition from SEM image patches via a CNN classifier. The confusion matrix gives the fraction of inclusions of each type that are classified with each predicted label; perfect accuracy would result in 1's down the diagonal. Sample image patches for each inclusion type are shown on the right.*

Finally, we note that the feature vector encodes the full range of visual information, some of which is not readily perceptible by human vision. For instance, chemical composition is not usually measured visually, but rather with specialized tools such as energy dispersive spectroscopy (EDS). However, chemical information is contained in the grayscale values of backscattered SEM images, albeit subtly. During steelmaking, EDS is used to determine the chemical composition of unintentional inclusions (slag, etc.), and SEM images of these inclusions are collected at the same time.[47] In order to determine whether SEM data alone contains sufficient information to determine inclusion composition, we utilized a pre-trained CNN[22, 28] and retrained the classifier using a data set of 2543 SEM inclusion images, balanced among five inclusion types. When tested on 509 previously unseen inclusion images, the system achieved 76% overall classification accuracy, considerably better than random chance accuracy of 20%, as shown in Figure 5(b). Furthermore, the confusion matrix confirms that the predominant misclassifications are among the sulfide inclusion types, which include significant compositional overlap. We note that these inclusions are virtually impossible to classify by eye (with the exception of the pores, which tend to be less circular than the other types). To perform this task, the CNN is presumably sensing inclusion shape, size, contrast, and color distribution with a fidelity that exceeds human perception, emphasizing the ability of CV and ML to augment and extend our ability to extract useful information from image data.

### 3.2 Semantic segmentation

Structural metrics, such as feature size, volume fraction, aspect ratio, etc., are the traditional quantities extracted from microstructural images using direct measurement tools. Typically, these measurements require a segmented image, where each pixel in the image is assigned to a microstructural constituent. Conventional automatic image segmentation algorithms, such as those incorporated in ImageJ,[48] generally operate by finding blobs of constant contrast or edges where contrast changes. While these approaches can work well on suitable microstructures, complex or non-ideal images often require considerable human intervention or even manual segmentation, resulting in a slow, material-specific, and subjective workflow.

Image segmentation is a foundational CV task, with important applications in robotics and medical imaging among others.[49] Thus, there is considerable research activity in developing segmentation methods. Because micrographs share features (e.g. edges, blobs, and visual textures) in common with natural images, we can adopt these methods to microstructural images via transfer learning. For example, we can use the PixelNet CNN[18b] trained on the ImageNet database of natural images[28] to compute a hypercolumn feature vector[18b] for each pixel in each micrograph. We then train PixelNet's pixel classification layers to classify pixels according to their microstructural constituent. Two examples are shown in Figure 6. In Figure 6(a) the system was trained using 20 hand-annotated images from the UHCS micrograph database,[50] and the results for one of the four previously-unseen test image are shown. In Figure 6(b) the system was trained on 30 hand-annotated images from a set of tomographic slices of an Al-Zn solidification dendrite,[51] again the results for one of the 10 previously-unseen image are given.

In both cases, segmentation accuracy is excellent: 93% for the steel microstructures and 99.6% for the Al-Zn. (Measures of precision and recall, which weight false positives and false negatives differently, do not indicate significant systematic errors, such as over- or under-prediction.) These segmentation maps are arguably equal in quality to the human annotations, and certainly adequate for quantitative analysis of the UHCS images (as demonstrated by DeCost et al.[50]) or tomographic reconstruction of the Al-Zn dendrite.[51] A significant benefit of this approach is that once the system is trained, subsequent image segmentations are calculated very quickly (near real time), and are autonomous, objective, and repeatable, enabling the high throughput necessary for applications such as 3D reconstruction or quality control.

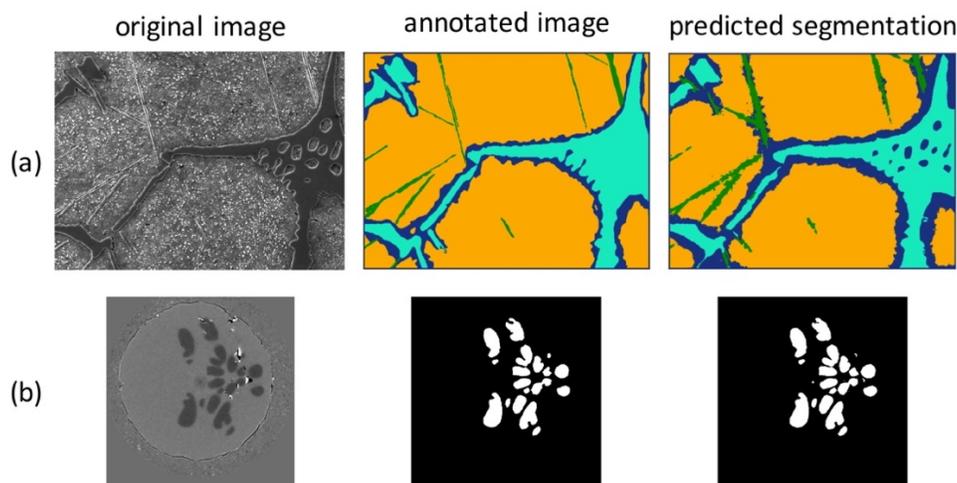

*Figure 6. Semantic segmentation of microstructural images using a pretrained CNN. (a) Semantic segmentation of microstructural constituents in SEM micrograph of ultrahigh carbon steel. Constituents include network carbide (light blue), ferritic denuded zone (dark blue), Widmanstatten carbide (green), and spheroidite matrix (gold). Note that the CNN segmentation captures "holes" in the network carbide that were unintentionally omitted by the human annotator. (b) Segmentation of a tomographic section of an Al-Zn alloy. The solidification dendrite is shown in white on a black background. Note that the CNN segmentation ignores prominent but irrelevant visual artifacts, including the sample edge, polishing defects, and the beam spot.*

An additional benefit of CNN-based image segmentation is the ability to capture human-like judgments about image features. For instance, in Figure 6(a), the spheroidite matrix constituent, comprised of spheroidite particles in a ferrite matrix, is segmented as a single constituent (orange). Conventional segmentation systems would be challenged to ignore the particles, which show up as distinct bright spots. Likewise, in Figure 6(b), the system learns to ignore sample preparation artifacts such as the sample edge, pores, and the circular beam spot at the center of the image. Again, these features are difficult to remove from conventional segmentation results. It is this capacity for learning what to look for and what to ignore that distinguishes the CV/ML approach to semantic segmentation.

Although the CV/ML system simulates some aspects of human visual judgment, it does not replicate human reasoning. Therefore, it is important to understand the strengths and limitations of CNN-based image segmentation in order to design the most effective tools. A small database of 17 SEM images (15 for training and 2 for testing) of nickel-based superalloy microstructures deformed in creep[52] provides an illustrative case study. These micrographs contain two prominent constituents: the oriented γ' cuboidal precipitates and dislocation lines [Figure 7(a)]. Both constituents are demarcated by narrow, linear features. Conventional image analysis [Figure 7(b)] is able to find these linear features, although it tends to overpredict them, creating many short, detached line segment artifacts that are not evident in the original image. A CNN-based image segmentation system (in this case, UNet[53]) can be trained to replicate the conventional analysis, and it has the advantage of being more resistant to line segment artifacts, as shown in Figure 7(c). However, the goal is to identify the dislocation segments only, ignoring the γ - γ' phase boundaries.

We initially attempted to modify the feature annotation training images in two ways: In one set, we manually eliminated the the γ - γ' phase boundaries, leaving only the dislocation annotations. However, the retrained system continued to capture many of the the γ - γ' boundaries. It was apparently unable to learn which of the linear features to ignore, at least for this small dataset. In a second set, we manually annotated γ - γ' phase boundaries as a separate constituent from dislocations. In this case, the system classified all near-horizontal and near-vertical linear features as γ - γ' boundaries, which resulted in an underprediction of dislocations. In both cases, the system was challenged to differentiate two microstructural constituents that are represented in the feature annotations as lines of single-pixel width.

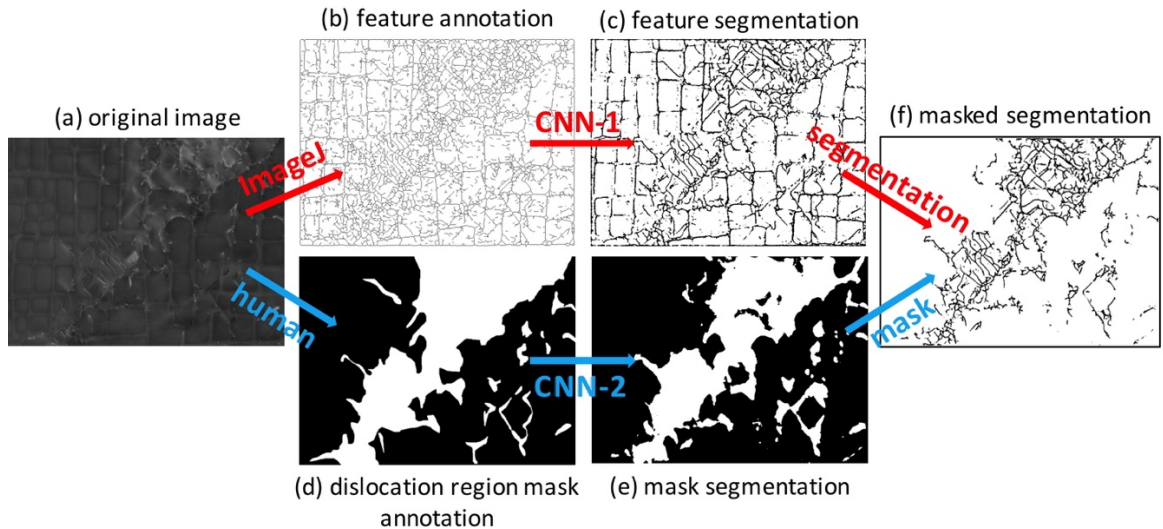

*Figure 7. Multipart segmentation of dislocation structures in SEM micrographs of nickel-based superalloys. (a) An original image showing the γ - γ' precipitates and dislocation traces. (b) ImageJ analysis annotates linear features. (c) CNN-based image segmentation finds linear features. (d) A human expert annotates the deformed and undeformed regions of the micrograph. (e) A second CNN-based image segmentation finds the dislocation mask. (f) Combining the feature segmentation (red arrows) and mask segmentation (blue arrows) yields a segmentation of the dislocation structure.*

High contrast gradients (lines or edges) are one kind of feature that the filter patches typically included in CNN architectures are able to sense. Other filter patches are optimized for uniform contrast areas (blobs), and still others for larger-scale visual textures (e.g. the lamellar structure of pearlite). Examining the superalloy micrographs, it becomes apparent that one way humans distinguish the dislocation lines from the γ-γ' boundaries is that the dislocations appear in regions of different visual texture. In fact, it is easy for a human to differentiate the deformed and undeformed regions of these micrographs. Using that insight, we manually annotated each image to segment these regions, as shown in Figure 7(d), and trained a separate CNN-based segmentation system to predict these dislocation region masks [Figure 7(e)]. By superposing the edge-based linear feature segmentation (red arrows in Figure 7) with the texture-based dislocation mask segmentation (blue arrows in Figure 7), we reconstitute an image segmentation that captures the dislocation lines and omits the γ-γ' boundaries. While the lack of an unambiguous ground truth makes it hard to evaluate the accuracy of these results, the quality of these segmentations was deemed sufficient to achieve the goal of comparing deformation states. As an added benefit, these segmentations are objective, repeatable, and self-consistent among images, so that the relative differences between images are quantitatively meaningful. Overall, achieving a useful segmentation on this challenging data set requires domain expertise in both microstructural images (materials science) and CNN-based segmentation systems (computational science). The payoff is the ability to expand the reach of traditional quantitative microstructure characterization to more complex microstructural features that have until now been difficult to treat in an automated fashion.

### 3.3 Object detection

Object detection entails locating each unique object of its kind in an image, i.e. finding each individual precipitate in a micrograph. When objects are spatially separated, object detection can be performed on a semantic segmentation using a conventional image analysis approach, such as a watershed algorithm[17f, 54] or connected-component labeling.[55] However, when objects overlap or occlude one another, as in the metal powder particles in Figure 8(a), automatic object detection becomes considerably more complex. Fortunately, object detection is another application of great interest in important CV applications such as self-driving vehicles, which must detect and account for each individual vehicle/pedestrian/tree in the field of view.[56]

Specialized CNNs, notably Faster R-CNN,[57] have been developed for object detection and the related task of instance segmentation. As in the case of semantic segmentation, discussed above, transfer learning allows these systems to be trained on natural images (such as the COCO detection data set[58]) and applied successfully to microstructural images. For example, in order to assess the prevalence of small satellite particles in a gas-atomized metal powder, the first task is to identify the discrete powder particles in a set of SEM images[59] [Figure 8(a)]. Tedious manual annotation yielded 5 images, each with several hundred powder particles outlined [Figure 8(b)]; three images were used to train a CNN instance segmentation system, and two were reserved to test the results in a cross-validation scheme. As shown in Figure 8(c), the system was successful at locating and delineating bounding boxes around individual particles, achieving an average match recall (correct particle identifications divided by all actual particles) of 80% and a match precision (correct particle identifications divided by the total number of particles identified) of about 94%. Loss of recall occurs when particles are missed, and unsurprisingly the system misses some small, irregularly shaped, and largely occluded particles. Although they comprise about 20% of the particles, they represent a very small fraction of the total particle volume. Loss of precision occurs when a particle is identified where none is present. However, the system is not finding particles where the image contains empty space. Instead, it is predicting that there are multiple particles where a single, agglomerated particle is identified in the manual annotation. In many of these cases, it is arguable whether the computer or human is correct. For this object detection task, the results of the CNN-based system are of good quality for subsequent analysis, and the errors that do occur are sensible and to some extent unavoidable.

Object detection is essential for calculating metrics based on counting objects, such as a number density or a population. However, in image analysis, it is more commonly a stepping stone to instance segmentation, as discussed in the next section.

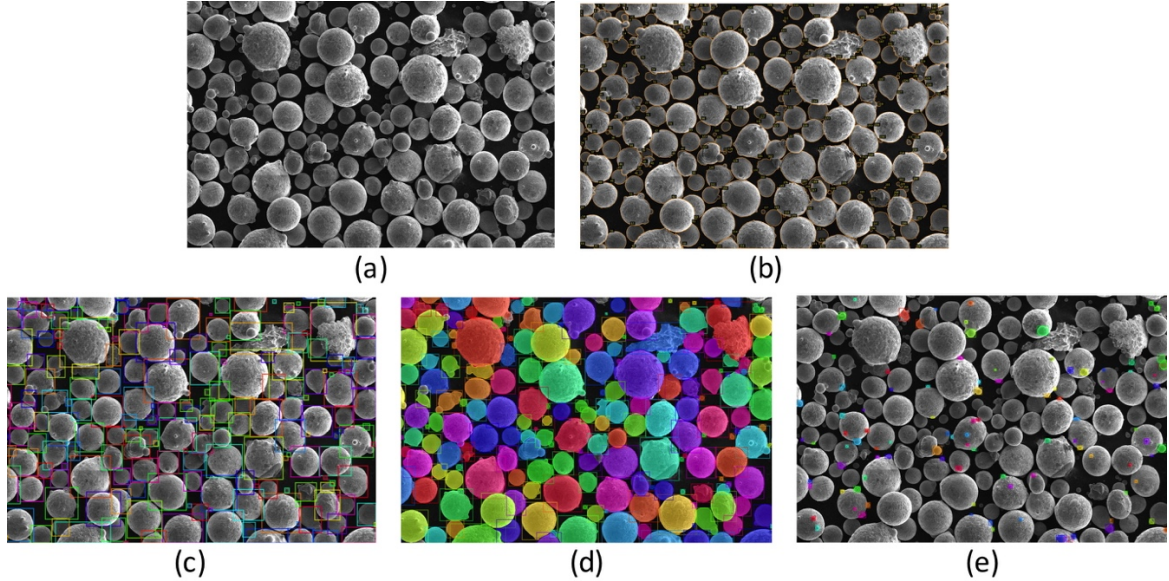

*Figure 8. Object identification and instance segmentation for SEM micrographs of gas atomized metal powders. (a) The original image contains dense and overlapping particles. (b) Manual annotation identifies and delineates individual particles in each image. (c) A CNN-based object identification system identifies particles and draws a bounding box around each one. (d) The same system segments (colors) each particle instance separately. (e) Following a similar workflow, satellite particles are identified and segmented. By overlaying on the particle segmentation, satellites can be associated with host particles.*

### 3.4  Instance segmentation

Once an object has been identified and bounded, instance segmentation is simply a matter of segmenting the pixels that belong to the object, as described above. In fact, object detection and instance segmentation are often combined in CNN implementations such as Mask R-CNN.[60] Figure 8(d) shows the instance segmentation, with each particle segmented (colored) as a separate object of particle type. The particle segmentations achieve a cross-validation average precision of 97.5% and recall of 95.4%; these high values indicate minimal over- or under-prediction and in fact are comparable to human performance, since there is some subjectivity in locating particle edges. While this application of instance segmentation found every object in the image, the same approach can differentiate and segment objects of particular types. Figure 8(e) shows instance segmentation of satellite particles, defined as small particles adhered to the surface of much larger particles. There is considerably more subjectivity in identifying satellites, which results in more disagreement between human and computer, thus lower match precision (69.2%) and recall (54.5%). However, this performance is still adequate to estimate satellite metrics such as satellite content as a function of particle size, and the autonomous nature of the CNN-based system allows much more throughput than would be feasible using manual segmentation.

### 4  NEXT STEPS: METRICS, APPLICATIONS, AND INTERPRETATION

### 4.1  Novel and advanced microstructural metrics

Beyond facilitating traditional microstructural measurements, CV image representations offer entirely new ways to characterize microstructures. For example, for a set of SEM micrographs of an Inconel-618 gas atomized metal powder[17f] we use connected-component labeling[55] to perform instance segmentation of each individual powder particle and then encode each particle patch via a CNN layer feature vector. After reducing the representation dimension via PCA,[29] we apply *k*-means unsupervised ML[36] to identify 8 visual clusters of particles. Figure 9(a) shows examples of particles belonging to each of the clusters. It is clear that this method is able to sort powder particles into visually similar groups. When the individual particle images are plotted in a t-SNE map [Figure 9(b)], particles cluster according to visual characteristics such as surface roughness, shape, and size. The statistics of the *k*-means clustering (i.e. particles per cluster) and the structure of the t-SNE plot (i.e. particle density map) act as fingerprints of the powder material, containing information about not only particle size, but also shape, roughness, agglomeration, etc. These representations can be applied to develop new, quantitative metrics to characterize powder materials that capture considerably more information that traditional powder size distributions.[61] We note that it is not feasible for humans to sort thousands of image patches, nor would we expect an objective and repeatable result. These new metrics are only accessible using a CV/ML approach, which leverages the ability of computers to perform repetitive tasks with the human-like visual judgment imparted by CV/ML methods.

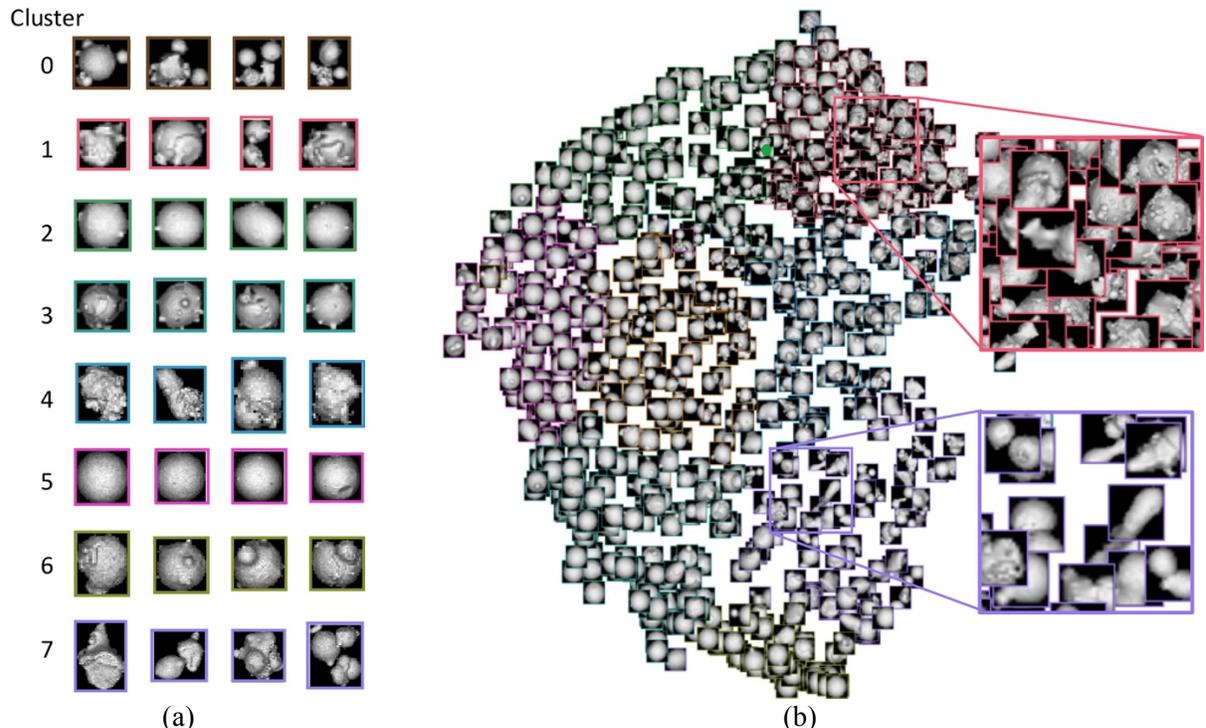

(a)  (b)

*Figure 9. The visual density map for a metal powder. (a) Individual particles are clustered visually; here an 8-cluster k-means analysis showing four particles per cluster. (b) A t-SNE visual similarity map for about 1100 Inconel-618 particles. Insets show how visually similar particles cluster (red = rough, agglomerated, purple = elongated). Box color corresponds to k-means cluster number. The k-means cluster statistics and image density map are quantitative fingerprints of the powder material.*

When image characterization is performed as a supervised task, it requires training and validation/test data that has been annotated with the ground truth (e.g. grain size, inclusion composition, microconstituent, etc.). Because annotation is often tedious and sometimes impossible, unsupervised image analysis is a goal in microstructural science and in CV more generally. The *k*-means clustering analysis described above suggests one route to an image metric (particle cluster statistics) using unsupervised ML. In this case, rather than designing an unsupervised ML method to measure a particular metric, we use the result of a standard, unsupervised ML method as the metric. Likewise, one can envision new microstructural metrics that arise directly from aspects of the CV/ML system, such as the feature vector or CNN filter activations.[62]

Finally, both microstructural science and CV share the goal of inferring 3D structure from 2D images.[63] Conventionally, this is the purview of quantitative stereology.[64] It currently remains an open question in CV/ML approaches for microstructural images.

## 4.2 CV/ML for processing- structure-property links

A primary objective of microstructural analysis is to discover processing-microstructure-property (PSP) relationships for systems of scientific and technological interest. While the application of CV/ML systems to PSP problems is beyond the scope of this overview, we will note that it is an active area of research. For instance, to make the processing/structure connection, CV/ML studies have examined micrograph databases to correlate microstructure with annealing history.[17b, 65] Likewise, CV/ML systems have been applied to relate microstructure with outcome properties including stress hot spot[66] and damage formation,[42] fatigue failure initiation,[67] fracture energy,[39a] ionic conductivity,[44] and fatigue strength.[65] Using the ability of CNNs to generate structures, several recent studies have also made strides toward the inverse problem of designing microstructures with target properties.[65, 68] A limiting factor in making robust PSP connects is the scarcity of large data sets that link specific microstructural images to processing history and/or property outcomes.

## 4.3 Interpretability: Opening the black box

Scientists and engineers can be hesitant to rely on "black box" algorithms, where the basis for a decision or prediction is unknown.[69] While this may not be a significant consideration for characterization tools such as semantic segmentation where performance can be assessed straightforwardly (and where the human version is not particularly interpretable either), it is critical for analysis tasks where understanding the basis for arriving at the conclusion is essential, such as making PSP connections. How black box CV systems make decisions is a significant, open question in computer science with many partial solutions, but none that are generally applicable across data sets and methods. Approaches include associating feature vector characteristics with microstructural length,[30] examining filter activations and characteristic textures,[30] and locating the image region most salient to decision-making.[70] In order to extract abstract scientific information from concrete visual information, the critical step is to identify the visual signatures of underlying physical processes. Discovering that signature in the feature vector how it is processed in the CV/ML pipeline is a grand challenge in AI-supported microstructural characterization and analysis.

## 5 CONCLUSIONS

The quantitative representation of microstructure is the foundational tool of microstructural science, connecting the materials structure to its composition, process history, and properties. Microstructural quantification traditionally involves a human deciding *a priori* what to measure and then devising a purpose-built method for doing so. However, recent advances in data science, including computer vision (CV) and machine learning (ML) offer new approaches to extracting information from microstructural images.

The key function of CV is to numerically encode the visual information contained in a microstructural image into a feature vector. The feature vector then provides input to ML algorithms that find associations and trends in the high-dimensional image representation. CV/ML systems for microstructural characterization and analysis span the taxonomy of image analysis tasks, including image classification, semantic segmentation, object detection, and instance segmentation. Applications include:

- Visual search, sort, and classification of micrographs via feature vector similarity.
- Extracting information not readily visible to humans, such as chemical composition in SEM micrographs, by using latent information in the feature vector.
- Performing semantic segmentation of microstructural constituents with a high accuracy and human-like judgment about what to look for and what to ignore.
- Combining segmentations based on different feature types to segment complex structures.

- Finding and bounding all instances of individual objects, even when they impinge and overlap.
- Segmenting individual objects to enable new capabilities in microstructural image analysis.

A common characteristic among all of these applications is that they capitalize on the ability of computational systems to produce accurate, autonomous, objective, repeatable results in an indefatigable and permanently available manner. These tools enable new approaches to microstructural analysis, including the development of new, rich visual metrics and the discovery of processing-microstructure-property relationships.

## 6 ACKNOWLEDGEMENTS

This work was supported by the National Science Foundation under grant CMMI-1826218 and the Air Force $D^3OM^2S$ Center of Excellence under agreement FA8650-19-2-5209. The authors appreciate data sets, collaborations, and helpful conversations provided by Dr. Iver Anderson (Ames National Laboratory), Dr. Brian DeCost (NIST), Anna Smith (Merck), Dr. Sabin Sulzer (Oxford), Prof. Peter Voorhees and Dr. Tiberiu Stan (Northwestern University), and Prof. Brian Webler (CMU).